\documentclass[conference]{IEEEtran}
\IEEEoverridecommandlockouts
\usepackage{cite}
\usepackage{amsmath,amssymb,amsfonts}
\usepackage{algorithmic}
\usepackage{graphicx}
\usepackage{textcomp}
\usepackage{xcolor}
\def\BibTeX{{\rm B\kern-.05em{\sc i\kern-.025em b}\kern-.08em
    T\kern-.1667em\lower.7ex\hbox{E}\kern-.125emX}}
\begin{document}

\title{TLCD: A Deep Transfer Learning Framework for Cross-Disciplinary Cognitive Diagnosis\\

\thanks{Corresponding author: Chunyan Zeng, Email: cyzeng@hbut.edu.cn.}
}

\author{\IEEEauthorblockN{Zhifeng Wang}
	\IEEEauthorblockA{\textit{CCNU Wollongong Joint Institute} \\
		\textit{Central China Normal University}\\
		Wuhan 430079, China \\
		zfwang@ccnu.edu.cn}
	\and
	\IEEEauthorblockN{Meixin Su}
	\IEEEauthorblockA{\textit{Faculty of Artificial Intelligence in Education} \\
		\textit{Central China Normal University}\\
		Wuhan 430079, China \\
		meixin\_su@mails.ccnu.edu.cn}
	\and
	\IEEEauthorblockN{Yang Yang}
	\IEEEauthorblockA{\textit{CCNU Wollongong Joint Institute} \\
		\textit{Central China Normal University}\\
		Wuhan 430079, China \\
		univeryang@ccnu.edu.cn}
	\and
	\IEEEauthorblockN{Chunyan Zeng}
	\IEEEauthorblockA{\textit{School of Electrical and Electronic Engineering} \\
		\textit{Hubei University of Technology}\\
		Wuhan 430068, China \\
		cyzeng@hbut.edu.cn}
	\and
	\IEEEauthorblockN{Lizhi Ye}
	\IEEEauthorblockA{\textit{Faculty of Artificial Intelligence in Education} \\
		\textit{Central China Normal University}\\
		Wuhan 430079, China \\		
		ye\_lizhi@ccnu.edu.cn}
}

%\author{\IEEEauthorblockN{1\textsuperscript{st} Meixin Su}
%\IEEEauthorblockA{\textit{Faculty of Artificial Intelligence in education} \\
%\textit{Central China Normal University}\\
%Wuhan, China \\
%meixin\underline{ }su@mails.ccnu.edu.cn}
%}

%\IEEEoverridecommandlockouts
%\IEEEpubid{\makebox[\columnwidth]{979-8-3315-1982-7/24/\$31.00 \copyright 2024 IEEE \hfill} 
%\hspace{\columnsep}\makebox[\columnwidth]{ }}

\maketitle

\IEEEpubidadjcol

\begin{abstract}
Driven by the dual principles of smart education and artificial intelligence technology, the online education model has rapidly emerged as an important component of the education industry. Cognitive diagnostic technology can utilize students' learning data and feedback information in educational evaluation to accurately assess their ability level at the knowledge level. However, while massive amounts of information provide abundant data resources, they also bring about complexity in feature extraction and scarcity of disciplinary data. In cross-disciplinary fields, traditional cognitive diagnostic methods still face many challenges. Given the differences in knowledge systems, cognitive structures, and data characteristics between different disciplines, this paper conducts in-depth research on neural network cognitive diagnosis and knowledge association neural network cognitive diagnosis, and proposes an innovative cross-disciplinary cognitive diagnosis method (TLCD). This method combines deep learning techniques and transfer learning strategies to enhance the performance of the model in the target discipline by utilizing the common features of the main discipline. The experimental results show that the cross-disciplinary cognitive diagnosis model based on deep learning performs better than the basic model in cross-disciplinary cognitive diagnosis tasks, and can more accurately evaluate students' learning situation.
\end{abstract}


\section{Introduction}
Cognitive diagnosis, as an important branch in the field of educational evaluation \cite{Li2026a,Wang2023j}, has gone through more than 30 years of development history since its birth, and more than 100 different models have emerged so far. The traditional cognitive diagnostic model mainly relies on linear psychological measurement functions to evaluate students' ability levels at the macro level \cite{Dong2023,Wang2025e,Li2023i,Wang2024b,Li2023g,Wang2024s,Li2023f,Wang2023d,Lyu2022}. Among them, the Item Response Theory (IRT) \cite{Lee2012,Wang2024p} and the Deterministic Inputs, Noise ``And'' Gate Model (DINA) \cite{DeLaTorre2009} are representative. The advantage of this type of model is that it has a solid theoretical foundation, can handle large-scale data well, and to some extent reveals the distribution characteristics of students' learning abilities. However, this traditional diagnostic method often cannot solve complex scenarios and multidimensional features, and relies heavily on manual annotation and expert intervention, making it difficult to fully capture students' complex cognitive processes and learning behaviors. In addition, traditional models are inadequate in handling high-dimensional data and complex relationships, which limits their application scope in modern educational evaluation \cite{Shi2026,Wang2024b,Wang2025b,Dong2025,Chen2025b,Liao2024,Chen2024e,Wang2023v,Ma2023b,Wang2022as}.

With the rapid development of technologies such as machine learning \cite{Zeng2018,Wang2025g,Zhu2013,Wang2018a,Tian2018,Wang2015b,Min2018,Wang2017,Zheng2025,Wang2015a,Zeng2025,Wang2011,Chen2025a,Wang2011a} and deep neural networks \cite{Zeng2025a,Li2023h,Zeng2024e,Wang2022ac,Zeng2023c,Wang2021,Zeng2022,Chen2025,Zeng2022b,Zheng2024,Zeng2021c,Wang2025f,Zeng2020a,Wang2025,Zeng2024g,Wang2025d,Zeng2024h,Wang2024m,Zeng2024b,Wang2023g,Zeng2024f,Wang2023v,Zeng2024c,Wang2023l,Zeng2024d,Wang2022at,Zeng2024,Chen2023b,Zeng2024a,Wang2023f,Zeng2023a,Wang2022t,Zeng2023,Wang2021m,Zeng2022a,Wang2020h,Zeng2021a,Zeng2021b,Zeng2020,Zeng2018}, more and more researchers are applying these emerging technologies to cognitive diagnostic modeling to analyze students' learning behavior and cognitive characteristics \cite{Liao2024}. These models make full use of big data and computing resources, and can not only handle high-dimensional and complex data relationships \cite{Li2023f,Wang2023j}, but also automatically learn and optimize model parameters, thereby improving the accuracy and efficiency of diagnosis \cite{Dong2024,Wang2023d,Chen2024g}. However, this field also faces some challenges and limitations. The complex model structure and high computational cost pose higher requirements for hardware equipment and algorithm optimization. At the same time, the interpretability of these models is relatively low, making it difficult for educators and students to directly understand and apply them. The effectiveness and reliability of emerging technologies in practical educational scenarios also need further validation and research.

In the new era of ``artificial intelligence education", accurately grasping students' knowledge structure and level in different subjects has become an urgent problem to be solved in the field of education. Previous cross-disciplinary cognitive diagnostic methods were often limited to individual diagnosis of independent disciplines and simple joint analysis. Although this approach considered the characteristics of disciplines to some extent, it failed to fully explore and utilize the inherent connections and mutual influences between disciplines.

Therefore, this paper introduces a cross-disciplinary cognitive diagnostic method based on deep learning, which achieves high-precision and efficient diagnosis and assessment of individuals' cognitive states in cross-disciplinary learning, providing strong support for personalized and targeted education. The primary contributions of this research are as follows:

1) Utilizing the deep learning technology's robust feature learning capabilities, we capture and extract complex patterns and regularities in individual learning processes. We also delve into the deep insights of students' learning behaviors, knowledge mastery, and strategies. 

2) To address the data disparities problem across disciplines, we employ transfer learning strategies, enhancing the model's performance in the target discipline through the main discipline's common features. This facilitates the efficient use of data resources across disciplines, reduces the cost of data acquisition and processing, and enhances diagnostic accuracy and generalizability. 

3) This paper conducted extensive experiments with a dataset from YNEG high school sophomores across 8 subjects, demonstrating the feasibility and effectiveness of the cross-disciplinary cognitive diagnostic method based on deep learning.

The rest of the paper is organized as follows. In Section \ref{RW}, we review the related work. In Section \ref{MED}, we introduce the main research methods. In Section \ref{EXP}, we describe the details of the experiment and the results. Finally, we have a summary of this work in Section \ref{CON}.

\section{Related Work} \label{RW}
The relevant work in this article can be summarized into two categories: cognitive diagnosis and transfer learning.
\subsection{Cognitive Diagnosis}
\subsubsection{Neural Network Cognitive Diagnostic Model}
NeuralCD (Neural Cognitive Diagnosis) \cite{Wang2023y} is a cognitive diagnosis framework based on neural networks, which models the complex interactive processes of students in problem-solving through neural networks. The input of NeuralCD is the one hot vector of the student and the one hot vector of the test question. When conducting cognitive diagnosis, the student's personal traits, relevant factors of the test question, and the interaction function of the model are fully considered to calculate the possibility of the student answering the test question correctly.

The student factor is designed to rate the student's proficiency in the knowledge concepts. NeuralCD \cite{Wang2023y} has been implemented in continuous form with the help of vector representation of the features in the DINA model. Specifically, NeuralCD describes a student by the vector ${F}^{s}$, where each element of the vector is continuous and indicates the student's proficiency in the knowledge concept. For example, if a student did a series of questions examining a total of 2 knowledge points, the student's ${F}^{s}= [0.9, 0.2]$ indicates that the student's mastery of the first knowledge point is relatively high at 0.9, but mastery of the second knowledge point is only 0.2. The student's knowledge proficiency vector is a parameter that was trained during the training process.

Test question factors are the characteristics of the test questions themselves. Test question factors can be categorized into knowledge point relevance vector ${F}^{kn}$. This vector represents the relationship between test questions and knowledge point concepts, and each element in the vector corresponds to a specific knowledge concept. The dimension of ${F}^{kn}$ is the same as the knowledge proficiency vector ${F}^{s}$, which indicates the relevance of the questions to the knowledge point. For example, a series of questions examined a total of 2 knowledge points, ${F}^{kn}= [0, 1]$ indicates that the question did not examine the 1st knowledge point and examined the 2nd knowledge point; (ii) other optional factors ${F}^{other}$. For example, the factors considered in IRT and DINA, such as the difficulty of the test questions, differentiation, and so on.

The NeuralCD framework skillfully utilizes the advantages of neural networks when constructing the interaction function between students and test questions. First, neural networks have powerful fitting ability, and neural networks in the NeuralCD framework are able to accurately construct the interaction model between students and test questions based on a variety of factors such as the students' historical performance, the difficulty of the questions, and the knowledge points. Second, neural networks have the ability to learn from limited data. In practical applications, educational data is often limited. Through the training and optimization of neural networks, the NeuralCD framework is able to distill useful information from limited data, and then construct accurate and stable interaction models.

\subsubsection{Neural Network Cognitive Diagnostic Model of Knowledge Relevance}

KaNCD (Knowledge-Association based Neural Cognitive Diagnosis) \cite{Wang2023y} is a neural network diagnostic method based on knowledge association. It is an extension of the NeuralCD model for assessing the cognitive state of a student during the learning process. KaNCD correlates a student's learning behaviors and performance with specific knowledge concepts (KC), and in turn diagnoses the student's proficiency in these concepts. The main advantage of KaNCD is that it is able to utilize a neural network to process a large amount of data and provide highly accurate and reliable diagnostic results.

In the KaNCD framework, the representations of students and knowledge concepts are not predetermined, but are gradually acquired through training and learning. At the beginning of training, the embedding vectors of students and knowledge concepts are randomly initialized. This initialization method ensures that the model is able to learn from various possible initial states during the training process, independent of the preset. During the training process, the model dynamically adjusts the values of the embedding vectors based on data such as students' learning behaviors and their performance in answering exercises, in order to better capture the relationship between students and knowledge concepts.

Specifically, each student (${S}_{i}$) and each knowledge concept (${K}_{j}$) are represented as vectors. During the training process, the A and B matrices are obtained by computing ${A}_{i,j}$ (the association matrix between student i and knowledge concept j) and ${B}_{i,j}$ (the confidence matrix between student i and knowledge concept j). Then, the training data is fed into the model to learn the association between students and knowledge points. This process can help KaNCD to better understand students' learning and knowledge acquisition, thus improving the accuracy and interpretability of the diagnosis.

\subsection{Transfer Learning}
The transfer learning aims to improve the performance of target domain tasks by utilizing source domain data or knowledge in cases where target domain data is scarce or labeling is difficult \cite{Wang2023f}. Transfer learning methods can be further subdivided into the following three categories:

\subsubsection{Instance-based transfer learning}

\begin{figure}[htbp]
\centerline{\includegraphics[scale=0.15]{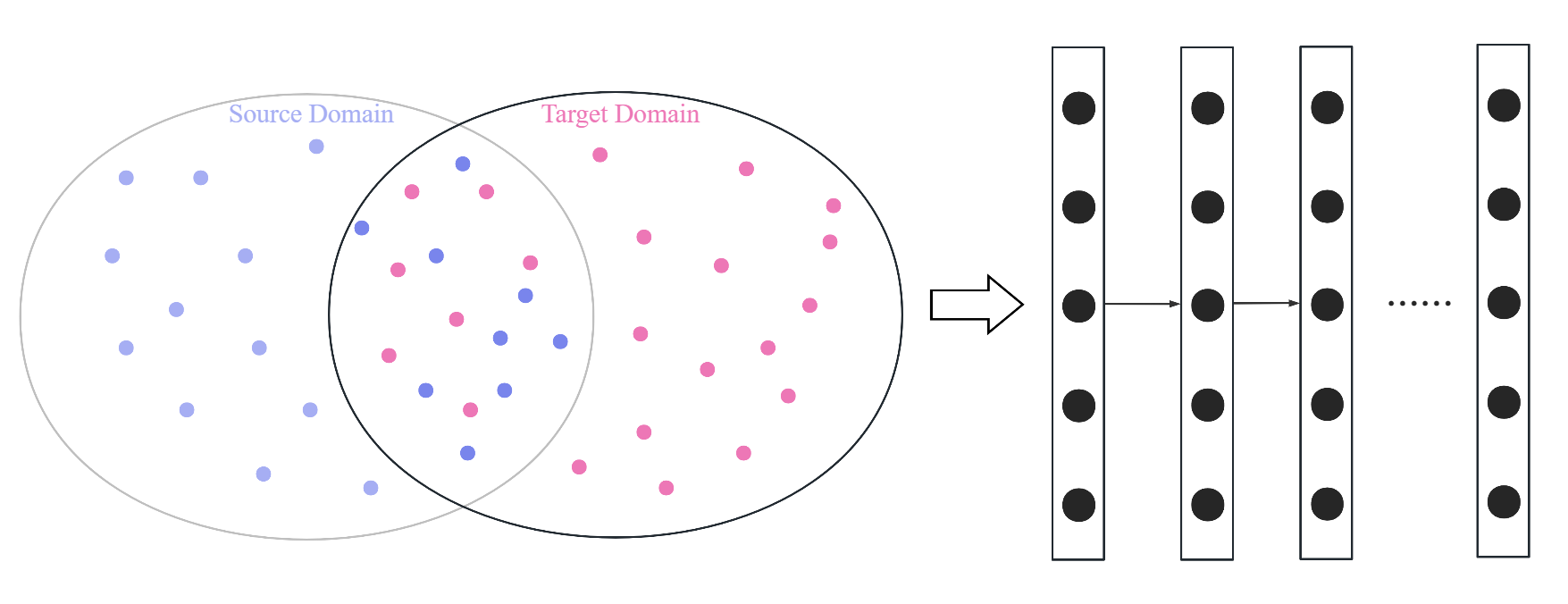}}
\caption{An instance-based approach to transfer learning.}
\label{fig1}
\end{figure}

The instance-based transfer learning method is a process of reusing data samples to different degrees according to specific weight generation rules, thus realizing transfer learning. Since in the process of transfer learning, not all data instances with markers in the source domain have practical application value to the target domain, as in Fig. \ref{fig1}. Therefore, the instance-based transfer learning method is committed to filtering out the instances that are beneficial to the target task and assigning appropriate weights according to their importance in the source domain. This personalized instance selection and weight assignment mechanism makes the model more flexible and robust, and can cope with the differences between different domains more effectively, thus achieving better transfer learning results.

\subsubsection{Feature-based transfer learning}

\begin{figure}[htbp]
\centerline{\includegraphics[scale=0.3]{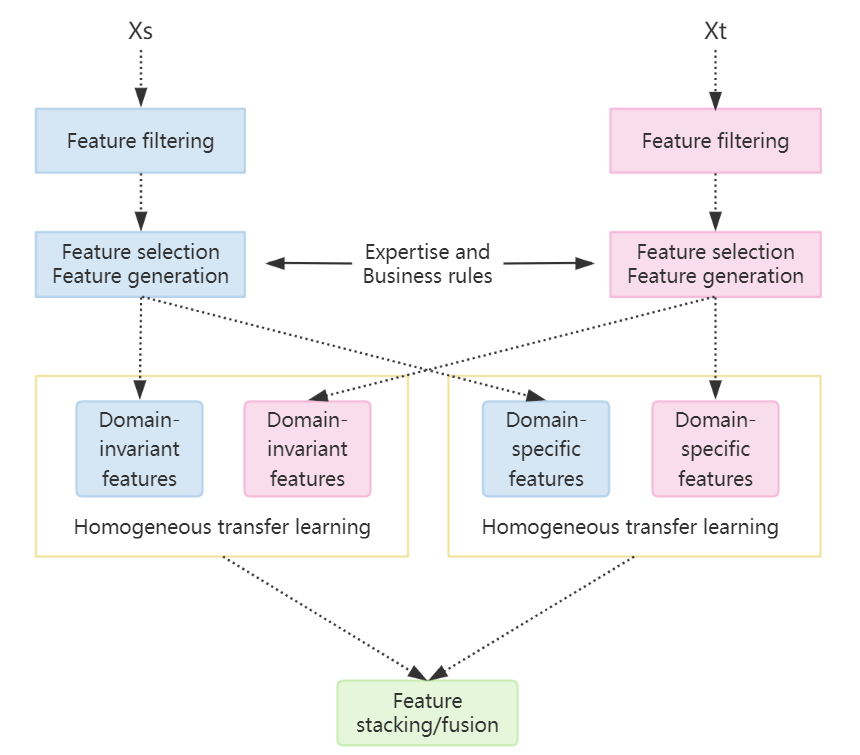}}
\caption{Isomorphic Transfer Learning vs. Heteromorphic Transfer Learning.}
\label{fig2}
\end{figure}

The feature-based transfer learning approach is a process of reusing shared features by filtering features that are valuable for the target task. This approach focuses on finding features that are shared and meaningful between the source and target domains for better transfer of knowledge and information between different domains, thus improving the generalization ability of the model on the target domain. As shown in Fig. \ref{fig2}, feature selection methods can reduce data dimensionality and improve the generalization ability of the model \cite{Li2023i}.

\subsubsection{Model-based transfer learning}

Model-based transfer learning relies on the correlation between the source and target domains and aims to transfer model parameters and feature representations from the source domain to the target domain to accelerate the convergence process of the model on the target task, as well as to improve its generalization ability to new data. Through the shared learning of model parameters and feature representations, the method enables the model to better adapt to the feature space of the target task, thus reducing the need for large-scale labeled data and improving the model's generalization ability and performance. Therefore, the model-based transfer learning approach is of great academic and applied significance in solving the problems of data scarcity and domain adaptation.

The pre-training and tuning parameter transfer learning method is a very popular model-based transfer learning strategy in the field of deep learning. In this approach, pre-training is first performed to learn the feature representation of the source task, and then fine-tuning is performed to adapt to the requirements of the target task to improve the performance performance. Through the combination of pre-training and fine-tuning, the model can converge faster and improve its generalization ability, greatly reducing the time and resource cost.

\section{Proposed Methods} \label{MED}

In this section, Neural Network Cognitive Diagnosis and Knowledge Associative Neural Network Cognitive Diagnosis are studied in depth, and a cross-disciplinary cognitive diagnosis method based on deep transfer learning is proposed. The method includes three segments: vector embedding, pre-training, and transfer learning.

\subsection{Cross-disciplinary cognitive diagnosis based on NeuralCD}\label{AA}

\begin{figure}[htbp]
\centerline{\includegraphics[scale=0.2]{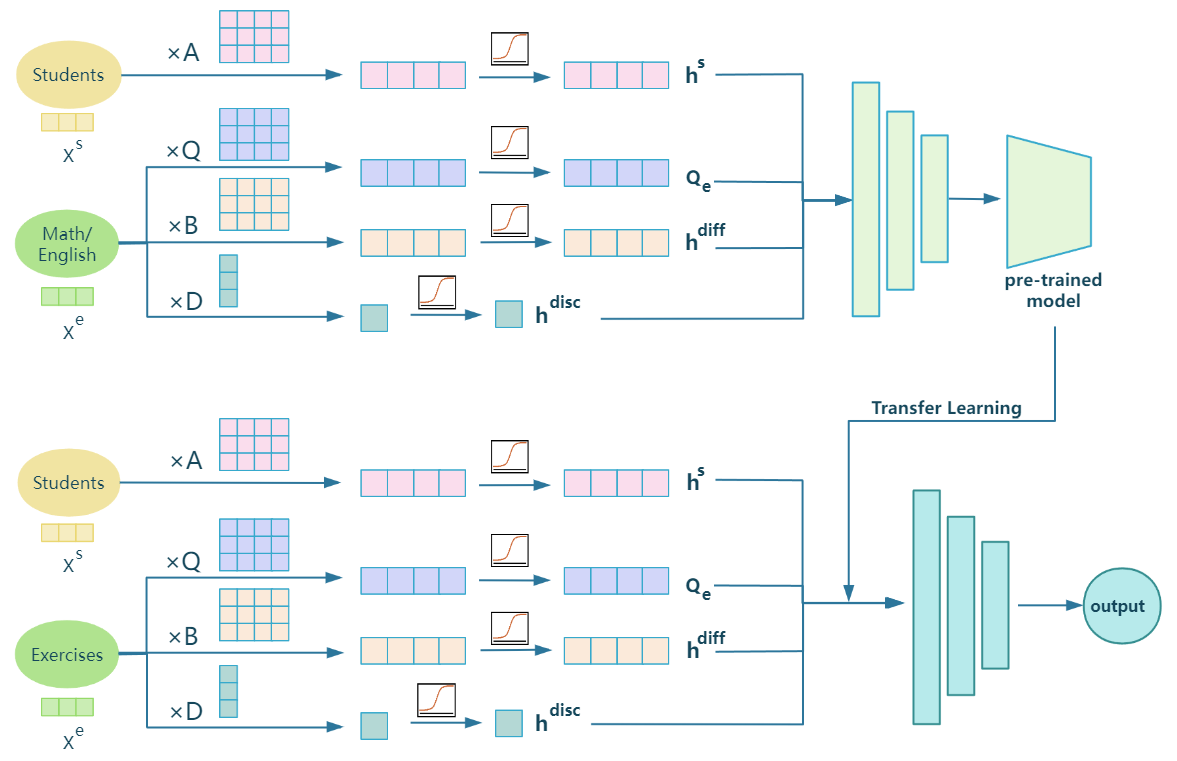}}
\caption{Cross-disciplinary cognitive diagnostic model.}
\label{fig3}
\end{figure}

Based on the NeuralCD model to introduce the transfer learning strategy, the neural network cognitive diagnostic model is used for pre-training, and the cross-disciplinary cognitive diagnostic model based on the NeuralCD model is constructed by adjusting the network as well as the parameters, as shown in Fig. \ref{fig3}. Overall, the model construction can be divided into vector embedding module, pre-training module, and transfer learning module.

\subsubsection{Vector Embedding}

This model performs cognitive diagnostic diagnosis by inputting the one-hot vectors of students and test questions from the answer records into a neural network. The student one-hot vector is denoted by ${x}^{s}\in {\{0,1\}}^{1\times N}$, and the one-hot vector of test questions is denoted by ${x}^{e}\in {\{0,1\}}^{1\times M}$. Each student's knowledge proficiency vector ${h}^{s}\in {\{0,1\}}^{1\times K}$ is obtained by multiplying ${x}^{s}$ with a trainable student proficiency matrix $A\in {R}^{N\times K}$; and $ Q= {\{Q}_{ij}\}_{M\times K}$ is the test question-knowledge The Q-matrix of correlation, ${Q}_{ij}=1$ means that KC $i$ examines knowledge point j, ${Q}_{ij}=0$ means that KC $i$ does not examine knowledge point j. The knowledge correlation vector, ${Q}_{e}\in {\{0,1\}}^{1\times K}$, is obtained directly from the Q-matrix.
\begin{equation}
{h}^{s}= sigmoid({x}^{s}\times A)\label{eq}
\end{equation}
\begin{equation}
{Q}_{e}= {x}^{e}\times Q\label{eq}
\end{equation}

In order to diagnose more accurately, two test factors were also used: the knowledge point difficulty vector of the test questions ${h}^{diff}\in{(0,1)}^{1\times K}$, which indicates the difficulty of the test questions in examining each knowledge point; and the differentiation vector of the test questions ${h}^{disc}\in(0,1)$, which indicates the test questions' ability to discriminate between students of different levels of proficiency.
\begin{equation}
{h}^{diff}= sigmoid({x}^{e}\times B),B\in {R}^{M\times K}\label{eq}
\end{equation}
\begin{equation}
{h}^{disc}= sigmoid({x}^{e}\times D),D\in {R}^{M\times 1}\label{eq}
\end{equation}

Where A, B, and D are trainable matrices.

Finally, the input layers of the multilayer neural network are as follows:
\begin{equation}
x= {Q}_{e}\circ ({h}^{s}-{h}^{diff})\times {h}^{disc}\label{eq}
\end{equation}

\subsubsection{Pre-training}

In the pre-training module, the student's knowledge proficiency in the subject is obtained by learning and training on the main subject and outputting the predicted score. The interaction function consists of a multilayer neural network that inputs the product of elements of the test question relevance vector and the student's proficiency vector, and passes through a number of fully connected layers and output layers to obtain the final prediction.

Two fully connected layers and one output layer are:
\begin{equation}
{f}_{1}= \phi ({W}_{1}\times{x}^{T}+{b}_{1})\label{eq}
\end{equation}
\begin{equation}
{f}_{2}= \phi ({W}_{2}\times{f}_{1}+{b}_{2})\label{eq}
\end{equation}
\begin{equation}
y= \phi ({W}_{3}\times{f}_{2}+{b}_{3})\label{eq}
\end{equation}

Where $\phi$ is the activation function sigmoid function, the sigmoid function is a logistic function that transforms any value between [0, 1] with the function expression:
\begin{equation}
sigmoid(x)= \frac{1}{1+{e}^{-x}}\label{eq}
\end{equation}

${x}^{T}$ is the transpose matrix of the input layer X; W and b are the weight matrix and bias vector of the cognitive state of each fully connected layer, respectively, which can be obtained from training.

%\begin{table}[htbp]
%\caption{Pre-trained model parameters based on NeuralCD transfer}
%\begin{center}
%\begin{tabular}{cc}
%\hline
%\textbf{Name} & \textbf{Value} \\
%\hline
%Batch size & 32 \\
%Layers num & 3 \\
%Learning rate & 0.002 \\
%\hline
%\end{tabular}
%\label{tab1}
%\end{center}
%\end{table}

\subsubsection{Transfer Learning}

In order to better utilize the features obtained from the pre-training, based on the idea of fine-tuning to obtain the part of the pre-trained model other than the output layer, the feature extraction layer as well as the weights of the original NeuralCDM model are frozen, two new fully-connected layers are added, and two Dropout layers are added, which are used to randomly throw away a certain proportion of the input features during the training process, in order to prevent overfitting.

In the new task, only the newly added fully connected layer is trained, passed to the output layer after processing by dropout rate=0.5, and finally the output is activated by the function.
\begin{equation}
{layer}_{1}=\phi ({W}_{4}\times {f}_{out})+{b}_{4}\label{eq}
\end{equation}
\begin{equation}
{layer}_{2}=\phi ({W}_{5}\times {layer}_{1})+{b}_{5}\label{eq}
\end{equation}
\begin{equation}
y=\phi ({W}_{6}\times {layer}_{2})+{b}_{6}\label{eq}
\end{equation}

This model is trained using cross entropy as a loss function to predict the binary cross entropy loss between the output value and the true value:
\begin{equation}
{loss}_{CDM}= -\sum ({r}_{i}log{y}_{i}+(1-{r}_{i})log(1-{y}_{i}))\label{eq}
\end{equation}

The proficiency vector ${h}^{s}$ is adjusted in the same direction as the change in the output prediction value y. At the end of the training, the student's corresponding ${h}^{s}$ is the diagnostic result for that student, and each dimension corresponds to that student's mastery on that knowledge point (range (0, 1)).

%\begin{table}[htbp]
%\caption{transfer model parameters based on NeuralCD transfer}
%\begin{center}
%\begin{tabular}{cc}
%\hline
%\textbf{Name} & \textbf{Value} \\
%\hline
%Batch size & 32 \\
%Layers num & 2 \\
%Newlayes num & 3 \\
%Learning rate & 1e-4\\
%\hline
%\end{tabular}
%\label{tab1}
%\end{center}
%\end{table}

\subsection{Cross-disciplinary cognitive diagnosis based on KaNCD}
\begin{figure}[htbp]
\centerline{\includegraphics[scale=0.2]{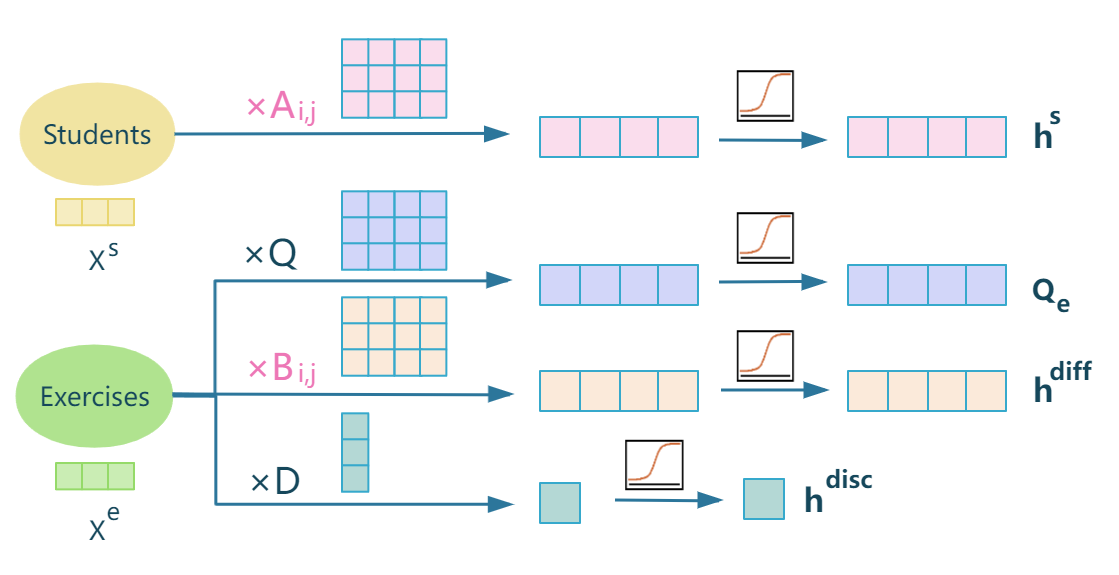}}
\caption{Vector embedding based on KaNCD transfer learning.}
\label{fig4}
\end{figure}

The model is based on transfer learning strategy and utilizes neural network cognitive diagnostic model for pre-training, as shown in Fig. \ref{fig4}. By adjusting the network and parameters, a cross-disciplinary cognitive diagnostic model based on KaNCD model is constructed. Overall, model construction can be divided into three modules: vector embedding, pre-training, and transfer learning.
\subsubsection{Vector Embedding}

This model inputs two types of data, namely student behavior data and knowledge point information, into a neural network and converts them into vector representations in high-dimensional space, thereby capturing the complex relationships between knowledge and knowledge, as well as between students. Behavioral data includes students' answering situations in a series of exercises, as well as the knowledge points involved in each exercise. Knowledge point information includes the attributes of each knowledge point, such as difficulty, distinguishability, etc.

Represent each student (${S}_{i}$) and each knowledge point (${K}_{j}$) as d-dimensional ($d<K$) potential vectors, i.e., ${l}_{i}^{s}$ and ${l}_{j}^{k}$. The ${l}_{i}^{s}$ vector represents the ith student and the ${l}_{j}^{k}$ vector represents the jth knowledge point. In a d-dimensional potential vector, each dimension represents a potential feature or attribute.

Through vector embedding, they capture information such as students' learning styles, abilities, and the difficulty and category of knowledge points. The researcher views the d-dimensional potential vectors as higher-order skills behind predefined knowledge concepts, and the value of each dimension in ${l}_{j}^{k}$ represents its preference for each higher-order skill.

The results obtained from the element-by-element multiplication drill of the two vectors ${l}_{i}^{s}$ and ${l}_{j}^{k}$ are then summed element-by-element, resulting in a vector that represents the student's proficiency on each knowledge concept. The proficiency of student ${S}_{i}$ on knowledge point ${K}_{j}$ is denoted as ${A}_{i,j}$, which is the weighted sum of the potential vectors of student ${S}_{i}$ and knowledge point ${K}_{j}$, and is calculated as follows:
\begin{equation}
{a}_{1}= {l}_{i}^{s}\circ{l}_{j}^{k}\label{eq}
\end{equation}
\begin{equation}
{a}_{i,j}= {W}_{a2}\times {a}_{i}+{b}_{a2}\label{eq}
\end{equation}

The difficulty of the test question ${e}_{i}$ on the knowledge point ${K}_{j}$ is denoted as ${B}_{i,j}$, which is the weighted sum of the potential vectors of the test question ${e}_{i}$ and the knowledge point ${K}_{j}$, as computed in the following formula:
\begin{equation}
{b}_{1}= {l}_{i}^{s}\circ{l}_{j}^{k}\label{eq}
\end{equation}
\begin{equation}
{b}_{i,j}= {W}_{b2}\times {b}_{i}+{b}_{b2}\label{eq}
\end{equation}

Where ${W}_{a2}$, ${b}_{a2}$, ${W}_{b2}$, and ${b}_{b2}$ are learnable parameters.

The knowledge proficiency vector ${h}^{s}$, the knowledge point relevance vector of the test questions ${Q}_{e}$, the knowledge point difficulty vector of the test questions ${h}^{diff}$, and the differentiation vector of the test questions ${h}^{disc}$ are as follows:

\begin{equation}
{h}^{s}= sigmoid({x}^{s}\times {A}_{i,j})\label{eq}
\end{equation}
\begin{equation}
{Q}_{e}= {x}^{e}\times Q\label{eq}
\end{equation}
\begin{equation}
{h}^{diff}= sigmoid({x}^{e}\times {B}_{i,j})\label{eq}
\end{equation}
\begin{equation}
{h}^{disc}= sigmoid({x}^{e}\times D),D\in {R}^{M\times 1}\label{eq}
\end{equation}

Among them, A, B, and D are trainable matrices.

Finally, the input layers of the multi-layer neural network are as follows:
\begin{equation}
x= {Q}_{e}\circ ({h}^{s}-{h}^{diff})\times {h}^{disc}\label{eq}
\end{equation}

\subsubsection{Pre-training}

Through the pre-training module, we use the main subject data (Math and English) to initialize the model so that the model can initially learn the intrinsic laws and structure of subject knowledge. This step not only improves the generalization ability of the model, but also provides strong support for subsequent transfer learning.

In order to be able to better fit the data or adapt to the task requirements, this network combines a matrix factorization (MF)-based predictive model and a neural network-based predictive model. The following parameters are accepted when initializing the network: the number of test questions M, the number of students N, the number of knowledge points K, the type of matrix decomposition T and the embedding dimension D.

With the idea of matrix decomposition, different computational methods such as simple inner product, generalized inner product, and different multilayer perceptual machines are selected according to different model type parameters T. Then, the students' proficiency in each knowledge point ${A}_{i,j}$ and the difficulty of each exercise corresponding to the knowledge point ${B}_{i,j}$ are calculated by means of inner product.

The elemental product of the test question relevance vector and the student proficiency vector $x= {Q}_{e}\circ ({h}^{s}-{h}^{diff})\times {h}^{disc}$ is fed into the neural network to generate the final prediction. The neural network consists of three fully connected layers:
\begin{equation}
{f}_{1}= \phi ({W}_{1}\times{x}^{T}+{b}_{1})\label{eq}
\end{equation}
\begin{equation}
{f}_{2}= \phi ({W}_{2}\times{f}_{1}+{b}_{2})\label{eq}
\end{equation}
\begin{equation}
y= \phi ({W}_{3}\times{f}_{2}+{b}_{3})\label{eq}
\end{equation}

Where $\phi$ is the activation function, ${x}^{T}$ is the transpose matrix of the input layer X; W and b are the weight matrix and bias vector of the cognitive state of each fully connected layer, respectively, which can be obtained from training.

%\begin{table}[htbp]
%\caption{Pre-trained model parameters based on KaNCD transfer}
%\begin{center}
%\begin{tabular}{cc}
%\hline
%\textbf{Name} & \textbf{Value} \\
%\hline
%Batch size & 32 \\
%Layers num & 3 \\
%Learning rate & 0.002 \\
%\hline
%\end{tabular}
%\label{tab1}
%\end{center}
%\end{table}

\subsubsection{Transfer Learning}

In order to apply the pre-trained model to a new cross-disciplinary cognitive diagnostic task, the following steps were taken to construct and adapt the model:

The parts of the pre-trained model other than the output layer containing the knowledge and feature representations learned by the model in pre-training were acquired. The feature extraction layers and the corresponding weights of the original KaNCD model were frozen to ensure that they remain unchanged during subsequent training.

Two new fully-connected layers are added on top of the frozen feature extraction layer to further extract and integrate features to generate a specific representation for the new task. Meanwhile, in order to enhance the generalization ability of the model, we add two Dropout layers between the fully-connected layers, which randomly discard a certain percentage of input features during the training process, helping to reduce the model's dependence on specific features.
\begin{equation}
{layer}_{1}=\phi ({W}_{4}\times {f}_{out})+{b}_{4}\label{eq}
\end{equation}
\begin{equation}
{layer}_{2}=\phi ({W}_{5}\times {layer}_{1})+{b}_{5}\label{eq}
\end{equation}
\begin{equation}
y=\phi ({W}_{6}\times {layer}_{2})+{b}_{6}\label{eq}
\end{equation}

Where $\phi$ is the activation function and ${f}_{out}$ is the output of the pre-training; W and b are the weight matrix and bias vector of the cognitive state of each fully connected layer, respectively.

%\begin{table}[htbp]
%\caption{transfer model parameters based on KaNCD transfer}
%\begin{center}
%\begin{tabular}{cc}
%\hline
%\textbf{Name} & \textbf{Value} \\
%\hline
%Batch size & 32 \\
%Layers num & 2 \\
%Newlayes num & 3 \\
%Learning rate & 0.002\\
%\hline
%\end{tabular}
%\label{tab1}
%\end{center}
%\end{table}

\section{Experimental Results and Analysis} \label{EXP}

In this section, the dataset, model performance, and case analysis are presented in detail. The designed experiment aims to answer the following questions:

\textbf{\textit{RQ1}}: How does the framework after transfer learning perform compared to the original cognitive diagnostic model?

\textbf{\textit{RQ2}}: How effective is this framework for predicting all disciplines?

\textbf{\textit{RQ3}}: Can this study improve the reliability of cross-disciplinary cognitive diagnostic models?

\subsection{Dataset}

\begin{table}[htbp]
\caption{YNEG Dataset Statistics and Related Information for Science}
\begin{center}
\begin{tabular}{ccccc}
\hline
\textbf{Data Statistics}&\multicolumn{4}{c}{\textbf{YNEG Science Dataset}} \\
\cline{2-4} 
\hline
Grade and Subject & \multicolumn{1}{l}{Math} & \multicolumn{1}{l}{Physics} & \multicolumn{1}{l}{Chemistry} & \multicolumn{1}{l}{Biology} \\
Data volume & 83584 & 43668 & 78960  & 87030  \\
Student numbers & 5224  & 3639 & 3290 & 2901 \\
Question numbers & 16 & 12& 24& 30\\
Knowledge numbers & 16 & 11  & 19 & 29 \\
\hline
\end{tabular}
\label{tab1}
\end{center}
\end{table}

\begin{table}[htbp]
\caption{YNEG Dataset Statistics and Related Information for Humanities}
\begin{center}
\begin{tabular}{ccccc}
\hline
\textbf{Data Statistics}&\multicolumn{4}{c}{\textbf{YNEG Humanities Dataset}} \\
\cline{2-4} 
\hline
Grade and Subject  & English  & History & Politics & Geography \\
Data volume  & 310245   & 38064   & 48432    & 67020\\
Student numbers  & 4773     & 1586    & 2018     & 2234\\
Question numbers  & 65       & 24      & 24       & 30 \\
Knowledge numbers & 23       & 24      & 24       & 14 \\
\hline
\end{tabular}
\label{tab2}
\end{center}
\end{table}

\begin{table}[htbp]
\caption{Description of some fields in YNEG dataset}
\begin{center}
\begin{tabular}{cc}
\hline
\textbf{Field Name} & \textbf{Meaning} \\
\hline
User id                             & Learner ID                   \\
Item id                             & Test ID                      \\
Score                               & Student answer scores        \\
Knowledge code                      & Knowledge point code         \\
Number of knowledge points (pieces) & 23  \\
\hline
\end{tabular}
\label{tab3}
\end{center}
\end{table}

This experiment uses the monthly exam answer dataset of YNEG high school sophomore students, with science subjects covering math, physics, chemistry, and biology, and humanities subjects covering English, history, politics, and geography. The data size and related information of the YNEG dataset are shown in Table \ref{tab1} and Table \ref{tab2}, and some field descriptions of the YNEG dataset are shown in Table \ref{tab3}.

\subsection{Model performance evaluation (\textit{RQ1}, \textit{RQ2})}
This experiment is based on the characteristics of the discipline and the scale of the data, and selects math and English as the main disciplines for pre-training. The experiment introduces transfer learning strategies based on the NeuralCD model and KaNCD model for cognitive diagnosis, predicting the probability of students answering questions correctly and their mastery of knowledge points. The predicted values are compared with the true values to obtain the AUC (Area Under ROC), ACC (Accuracy), RMSE, and MAE.

\begin{table}[htbp]
\caption{Performance of the NeuralCD model on datasets}
\begin{center}
\begin{tabular}{ccccc}
\hline
\textbf{Discipline}&\multicolumn{4}{c}{\textbf{Index}} \\
\cline{2-5} 
                            & \textbf{AUC(\%)} & \textbf{ACC(\%)} & \textbf{RMSE} & \textbf{MAE} \\
\hline
Math                 & 81.5062     & 75.1765     & 0.412614      & 0.351146     \\
Physics                     & 76.8717     & 73.7405     & 0.423814      & 0.358252     \\
Chemistry                   & 65.2354    & 63.7200     & 0.472712      & 0.441762     \\
Biology                     & 70.6790     & 67.9792     & 0.449756      & 0.414787     \\
English                     & 73.7804     & 67.9330     & 0.454552      & 0.401371     \\
History                     & 72.4226     & 65.2128     & 0.470262      & 0.447815     \\
Politics                    & 71.3724     & 67.1025     & 0.456210      & 0.423143     \\
Geography                   & 71.6398     & 66.0698    & 0.460095      & 0.427174    \\
\hline
\end{tabular}
\label{tab4}
\end{center}
\end{table}

\begin{table}[htbp]
\caption{Performance on datasets after transfer learning based on NeuralCD models}
\begin{center}
\begin{tabular}{ccccc}
\hline
\textbf{Discipline}&\multicolumn{4}{c}{\textbf{Index}} \\
\cline{2-5} 
                            & \textbf{AUC(\%)} & \textbf{ACC(\%)} & \textbf{RMSE} & \textbf{MAE} \\
\hline
Math                 & /            & /            & /             & /            \\
Physics                     & 76.9601     & 73.8819     & 0.423220      & 0.314988     \\
Chemistry                   & 65.2993     & 62.8673     & 0.474195      & 0.439652     \\
Biology                     & 70.7520     & 69.2202     & 0.446213      & 0.385677     \\
English                     & /            & /            & /             & /            \\
History                     & 72.4307     & 67.1046     & 0.460934      & 0.431876     \\
Politics                    & 71.4573     & 67.3503     & 0.455189      & 0.421094     \\
Geography                   & 71.6616     & 66.5075     & 0.458471      & 0.421665    \\
\hline
\end{tabular}
\label{tab5}
\end{center}
\end{table}

Focusing on the performance on physical subjects after transfer learning based on NeuralCD model, as in Table \ref{tab4} and \ref{tab5} the AUC is improved by 0.9\% compared to unmigrated; the AUC is improved by 0.14\% compared to unmigrated; and the MAE is improved by a reduction of 4.4\% compared to unmigrated. Even better results are shown in terms of predictive ability.

Compared with NeuralCD, a neural network cognitive diagnostic model without the introduction of the transfer learning strategy, the model shows some improvement in the performance of AUC, ACC, MAE, and RMSE, which proves that cross-disciplinary diagnosis based on the transfer of the NeuralCD model can effectively improve the accuracy of cognitive diagnosis prediction.

\begin{table}[htbp]
\caption{Performance of the KaNCD model on datasets.}
\begin{center}
\begin{tabular}{ccccc}
\hline
\textbf{Discipline}&\multicolumn{4}{c}{\textbf{Index}} \\
\cline{2-5} 
                            & \textbf{AUC(\%)} & \textbf{ACC(\%)} & \textbf{RMSE} & \textbf{MAE} \\
\hline
Math                 & 84.8834     & 77.6023     & 0.390532      & 0.302107     \\
Physics                     & 81.3209     & 76.8702     & 0.405063      & 0.344100     \\
Chemistry                   & 70.9586     & 67.5701     & 0.457287      & 0.412084     \\
Biology                     & 72.4932     & 70.3692     & 0.440715      & 0.395685     \\
English                     & 78.3413     & 71.0166     & 0.434569      & 0.378053     \\
History                     & 73.3159     & 67.7352     & 0.462426      & 0.423773     \\
Politics                    & 74.4810     & 67.4467     & 0.458068      & 0.385257     \\
Geography                   & 72.6718     & 66.8656     & 0.454737      & 0.413167      \\
\hline
\end{tabular}
\label{tab6}
\end{center}
\end{table}

\begin{table}[htbp]
\caption{Performance on datasets after transfer learning based on KaNCD models}
\begin{center}
\begin{tabular}{ccccc}
\hline
\textbf{Discipline}&\multicolumn{4}{c}{\textbf{Index}} \\
\cline{2-5} 
                            & \textbf{AUC(\%)} & \textbf{ACC(\%)} & \textbf{RMSE} & \textbf{MAE} \\
\hline
Math                          & /           & /           & /             & /            \\
Physics                              & 81.8235     & 77.0840     & 0.405482      & 0.300570     \\
Chemistry                            & 71.1451     & 68.3384     & 0.456676      & 0.399385     \\
Biology                              & 76.7199     & 71.9933     & 0.426201      & 0.353066     \\
English                              & /           & /           & /             & /            \\
History                              & 73.9314     & 67.9453     & 0.457185      & 0.407301     \\
Politics                             & 77.2760     & 70.4198     & 0.437858      & 0.389458     \\
Geography                            & 74.6668     & 68.5069     & 0.446990      & 0.397442    \\
\hline
\end{tabular}
\label{tab7}
\end{center}
\end{table}

As in Table \ref{tab6} and \ref{tab7}, comparing the performance of the model in physics before and after transfer: the AUC score reached 74.3210\% without transfer and 77.2760\% after transfer; The ACC score did not transfer to 68.0523\%, but after transfer, it reached 0.704198; The RMSE score reached 0.455649 without transfer and 0.437858 after transfer; The MAE score did not migrate to 0.428689, but after transfer, it reached 0.389458.

Focusing on humanities disciplines, the AUC of transfer learning based on the KaNCD model showed a significant improvement compared to non-transfer, while the AUC value of political disciplines increased by 3\%, demonstrating superior predictive ability. This indicates that the model after transfer learning can achieve better generalization performance in cross-disciplinary cognitive diagnostic tasks.

\subsection{Case analysis (\textit{RQ3})}
In terms of cross-disciplinary knowledge cognitive diagnosis, we selected a student for case analysis and conducted an in-depth analysis of their scores in various subjects in the optimal epoch. Using a transfer learning model optimized based on the NeuralCD model and a transfer learning model optimized based on KaNCD, predict the answering situation of the student in eight subjects. Among them, the blue dots represent the true values, and the green dots represent the predicted values.

\begin{figure}[htbp]
\centerline{\includegraphics[scale=0.5]{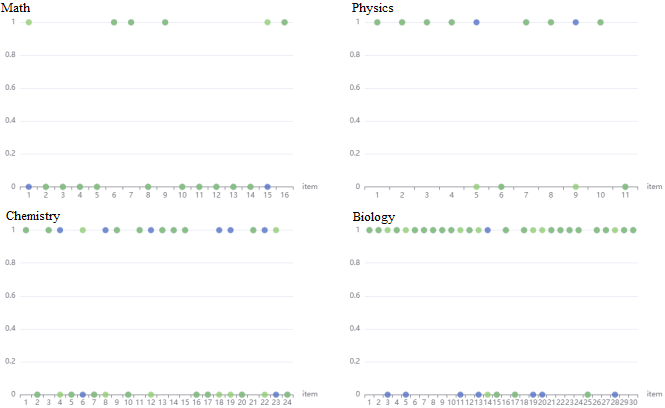}}
\caption{Scatter plot of transfer learning model optimized based on NeuralCD model for predicting answers in science subjects.}
\label{fig5}
\end{figure}

In Fig. \ref{fig5}, green dots indicate correct predictions (predicted values cover true values), and blue dots indicate incorrect predictions (true values are not covered), indicating excellent predictive performance of the model. Among the four science disciplines, the accuracy rate of math is 87.5\%, physics is 83.33\%, chemistry is 66.67\%, and biology is 70\%.

\begin{figure}[htbp]
\centerline{\includegraphics[scale=0.5]{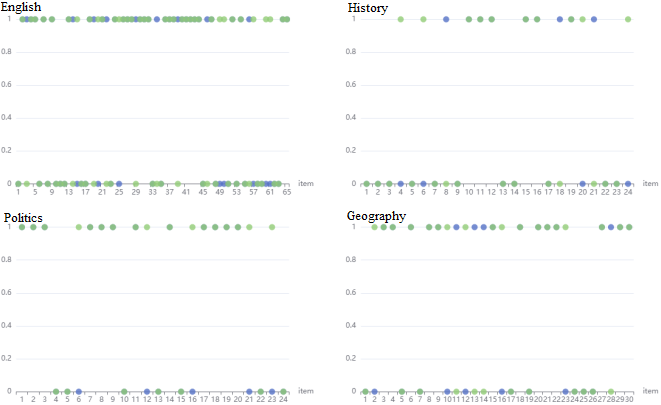}}
\caption{Scatter plot of transfer learning model optimized based on NeuralCD model for predicting answers in humanities subjects.}
\label{fig6}
\end{figure}

In Fig. \ref{fig6}, green dots indicate correct predictions (predicted values cover true values), and blue dots indicate incorrect predictions (true values are not covered). It can be seen that the model's performance in predicting humanities disciplines is also relatively good. The accuracy rate of English subject is 73.85\%, the accuracy rate of history subject is 70.83\%, the accuracy rate of political subject is 79.17\%, and the accuracy rate of geography subject is 70\%.

By drawing a scatter plot, we can visually see the student's scores in different subjects. Such score prediction analysis helps teachers to have a more comprehensive understanding of students' learning status, quickly identify students' strengths and weaknesses, develop targeted teaching plans based on different students' situations, help students overcome learning difficulties, and improve learning outcomes.

\begin{figure}[htbp]
\centerline{\includegraphics[scale=0.5]{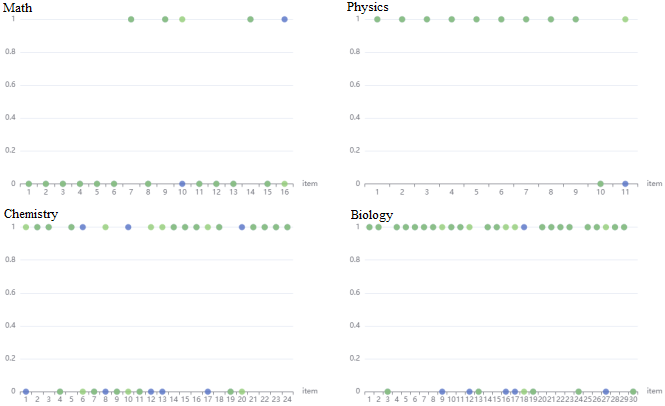}}
\caption{Scatter plot of transfer learning model optimized based on KaNCD model for predicting answers in science subjects.}
\label{fig7}
\end{figure}

In Fig. \ref{fig7}, the green dots indicate correct predictions (the predicted values cover the true values), and the blue dots indicate incorrect predictions (the true values are not covered), indicating excellent performance of the model's prediction. The accuracy rate of math is 87.5\%, physics is 91.67\%, chemistry is 66.67\%, and biology is 80\%.

\begin{figure}[htbp]
\centerline{\includegraphics[scale=0.5]{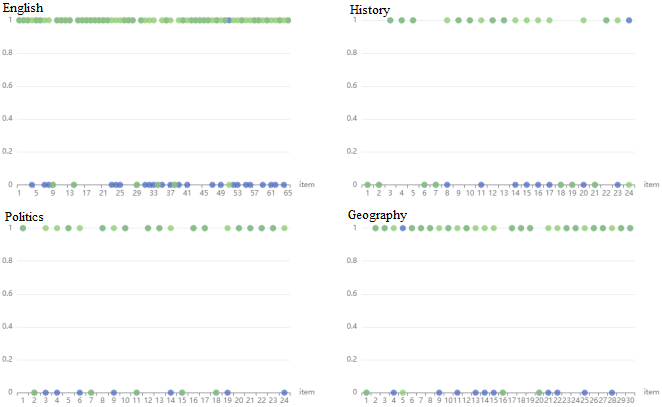}}
\caption{Scatter plot of transfer learning model optimized based on KaNCD model for predicting answers in humanities subjects.}
\label{fig8}
\end{figure}

In Fig. \ref{fig8}, green dots indicate correct predictions (predicted values cover true values), and blue dots indicate incorrect predictions (true values are not covered), indicating excellent predictive performance of the model. The accuracy rate of English subject is 63.0769\%, the accuracy rate of history subject is 62.5\%, the accuracy rate of political subject is 70.83\%, and the accuracy rate of political subject is 63.33\%.

\section{Conclusions} \label{CON}
This paper is based on a cognitive diagnosis model using neural networks and knowledge association neural networks, and takes the monthly exam answer dataset of YENG high school sophomore students as the research object to deeply explore cross-disciplinary cognitive diagnosis methods. This study introduces transfer learning strategies aimed at optimizing the performance of NeuralCD and KaNCD models in cross-disciplinary cognitive diagnosis, achieving their application on the YNEG dataset and effectively completing a comprehensive diagnosis of students' knowledge mastery in different disciplines. Future research can attempt to introduce more transfer learning strategies and compare the impact of different strategies on model performance in order to find the optimal transfer method.

\bibliographystyle{IEEEtran}
\bibliography{Ref,Citations}

\end{document}